\pdfoutput=1

\documentclass[11pt]{article}

\usepackage{acl}

\usepackage{times}
\usepackage{latexsym}
\usepackage{amsmath}
\usepackage{amssymb}
\usepackage{subfigure}
\usepackage{multirow}
\usepackage{graphicx}
\usepackage{tikz}
\usepackage{pgfplots}
\usepackage{booktabs}
\usepackage{csvsimple}
\usepgfplotslibrary{groupplots}
\usepackage[T1]{fontenc}

\usepackage[utf8]{inputenc}

\usepackage{microtype}

\usepackage{cleveref}
\crefformat{section}{\S#2#1#3} 
\crefformat{subsection}{\S#2#1#3}
\crefformat{subsubsection}{\S#2#1#3}

\title{Hybrid Transducer and Attention based Encoder-Decoder Modeling for Speech-to-Text Tasks}

\author{Yun Tang${}^\star$, Anna Y. Sun${}^\star$, Hirofumi Inaguma${}^\star$, Xinyue Chen${}^\diamondsuit$\thanks{ $\,\,\,$Xinyue Chen contributed to this work during her internship at Meta.}, 
\\ \textbf{Ning Dong${}^\star$, Xutai Ma${}^\star$, Paden D. Tomasello${}^\star$, Juan Pino${}^\star$} \\
 \\
   Meta AI${}^\star$, $\,\,\,\,$Carnegie Mellon University${}^\diamondsuit$ \\
  \texttt{\{yuntang,hirofumii\}@meta.com} 
}

\begin{document}
\maketitle
\begin{abstract}
Transducer and Attention based Encoder-Decoder (AED) are two widely used frameworks for speech-to-text tasks. They are designed for different purposes and each has its own benefits and drawbacks for speech-to-text tasks.
In order to leverage strengths of both modeling methods, we propose a solution by combining Transducer and Attention based Encoder-Decoder (TAED) for speech-to-text tasks. 
The new method leverages AED's strength in non-monotonic sequence to sequence learning while retaining Transducer's streaming property.
In the proposed framework, Transducer and AED  
 share the same speech encoder. The predictor in  Transducer is replaced by the decoder in 
the AED model, and the outputs of the decoder are conditioned on the speech inputs instead of outputs from an unconditioned language model. 
The proposed solution ensures that the model is optimized by covering all possible read/write scenarios and creates a matched environment for streaming applications. 
We evaluate the proposed approach on the \textsc{MuST-C} dataset and the findings demonstrate that TAED  performs significantly better than Transducer for offline automatic speech recognition (ASR) and speech-to-text translation (ST) tasks. In the streaming case, TAED outperforms Transducer in the ASR task and one ST direction while comparable results are achieved in another translation direction.
\end{abstract}
\section{Introduction}
Neural based end-to-end frameworks have achieved remarkable success in speech-to-text tasks, such as automatic speech recognition (ASR) and speech-to-text translation (ST)~\citep{Li2021RecentAI}. 
These frameworks include Attention based Encoder-Decoder modeling (AED)~\citep{Bahdanau2014NeuralMT}, connectionist temporal classification (CTC)~\citep{Graves2006ConnectionistTC} and Transducer~\citep{Graves2012SequenceTW} etc.
They are designed with different purposes and have quite different behaviors, even though all of them could be used to 
solve the mapping problem from a speech input sequence to a text output sequence. 

AED handles the sequence-to-sequence learning by allowing the decoder to attend to parts of the source sequence. It provides a powerful and general solution that is not bound to the input/output modalities, 
lengths, or sequence orders. Hence, it is widely used for  ASR~\citep{Chorowski2015AttentionBasedMF,Chan2015ListenAA,Zhang2020TransformerTA,Gulati2020ConformerCT,Tang2020AGM},  
and ST~\citep{Berard2016ListenAT,Weiss2017SequencetoSequenceMC,Li2021MultilingualST,Tang2022UnifiedSP}. 

CTC and its variant Transducer are designed to handle monotonic alignment between the speech input sequence and text output sequence. 
A hard alignment is generated between speech features and target text tokens during decoding, in which every output token is associated or synchronized with an input speech feature.
CTC and Transducer have many desired properties for ASR. For example, they fit into streaming applications naturally, and the 
input-synchronous decoding can help alleviate over-generation or under-generation issues within AED. ~\citet{Sainath2019TwoPassES,Chiu2019ACO} show that Transducer achieves better WER than AED in long utterance recognition, while AED outperforms Transducer in the short utterance case. 
On the other hand, CTC and Transducer are shown to be suboptimal in dealing with non-monotonic sequence mapping~\citep{Chuang2021InvestigatingTR}, though some initial attempts show encouraging progress~\citep{Xue2022LargeScaleSE,Wang2022LAMASSUSL}. 

In this work, we propose a hybrid Transducer and AED model (TAED), which integrates both AED and Transducer models into one framework 
to leverage strengths from both modeling methods.
In TAED, we share the speech encoder between AED and Transducer. The predictor in Transducer is replaced with the decoder in AED.  
The AED decoder output assists the Transducer's joiner to predict the output tokens. 
Transducer and AED models are treated equally and optimized jointly during training, while only Transducer's joiner outputs are used during inference.
We extend the TAED model to streaming applications under the chunk-based synchronization scheme, which guarantees full coverage of read/write choices in the training set and removes the training and inference discrepancy.
The relationship between streaming latency and AED alignment is studied, and a simple, fast AED alignment is proposed to achieve low latency with small quality degradation.
The new approach is evaluated in ASR and ST tasks for offline and streaming settings.
The results show the new method helps to achieve new state-of-the-art results on offline evaluation.
The corresponding streaming extension also improves the quality significantly under a similar latency budget.
To summarize, our contributions are below:
\begin{enumerate}
    \item TAED, the hybrid of Transducer and AED modeling, is proposed for speech-to-text tasks
    \item A chunk-based streaming synchronization scheme is adopted to remove the training and inference discrepancy for streaming applications
    \item A simple, fast AED alignment is employed to balance TAED latency and quality 
    \item The proposed method achieves SOTA results on both offline and streaming settings for ASR and ST tasks
\end{enumerate}




\begin{figure*}
\hfill
\centering
\subfigure[ Attention based encoder decoder.]{\includegraphics[width=0.50\columnwidth]{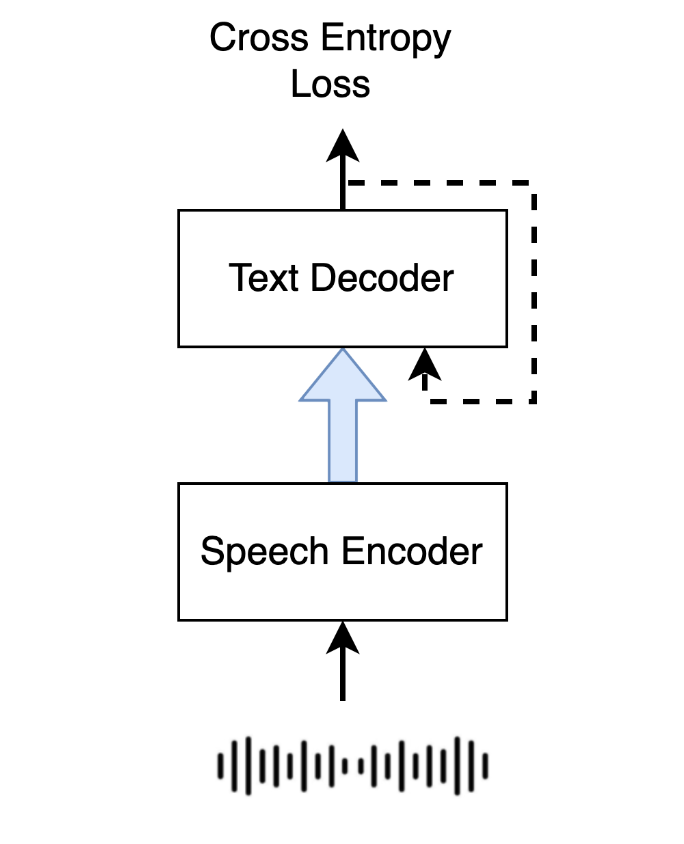}\label{fig:ead}}
\hfill
\subfigure[ Transducer]{\includegraphics[width=0.65\columnwidth]{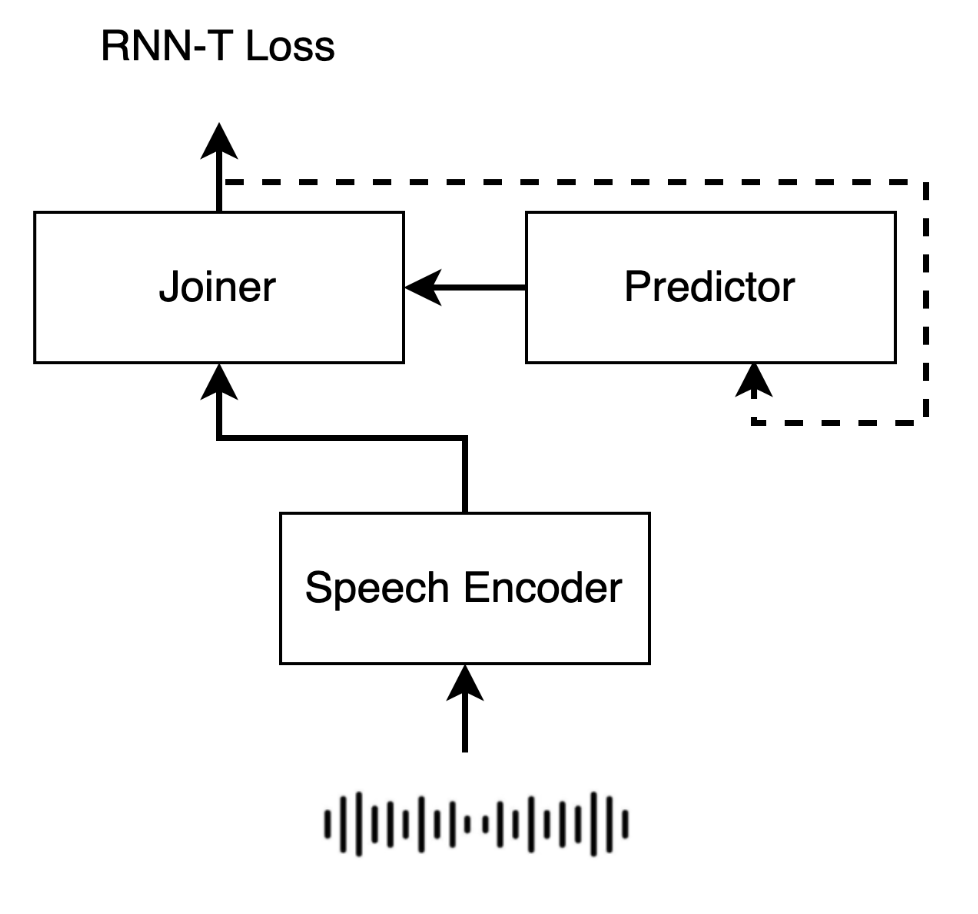}\label{fig:transducer}}
\hfill
\subfigure[ Hybrid transducer and attention based encoder-decoder.]{\includegraphics[width=0.65\columnwidth]{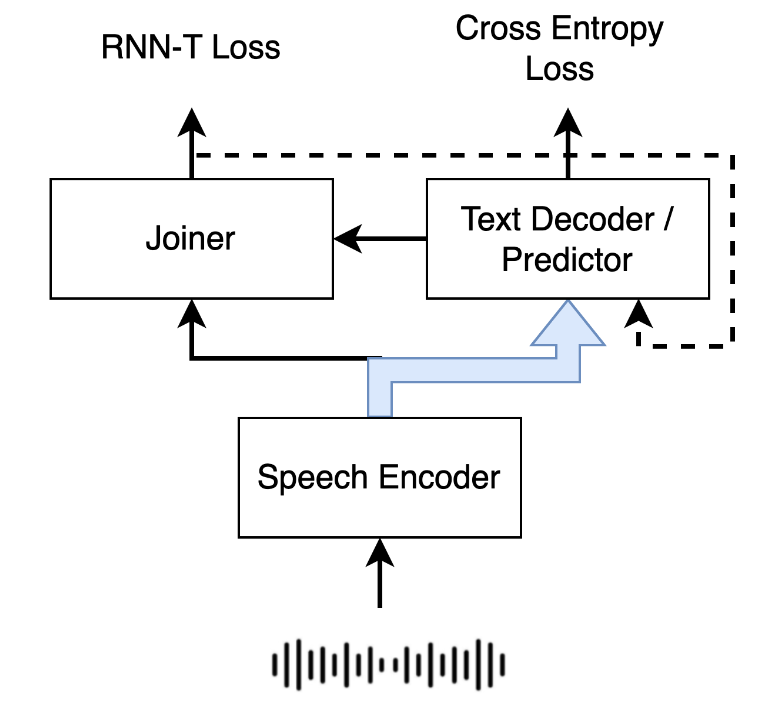}\label{fig:hybrid_TAED}}
\hfill
\caption{Comparison among different frameworks: AED, Transducer and TAED.}
\label{fig:framework_compare}
\end{figure*}

\section{Preliminary}
Formally, we denote a speech-to-text task training sample as a $(\mathbf{x}, \mathbf{y})$ pair. $\mathbf{x}=x_{1:T}$ and $\mathbf{y}=y_{1:U}$ are the speech input features and target text tokens, respectively. 
$T$ and $U$ are the corresponding sequence lengths. $y_u \in \mathcal{V}$ and  $\mathcal{V}$ is the target vocabulary.   The objective function is to minimize the negative log likelihood  $\log p(\mathbf{y}|\mathbf{x},\theta)$  over the training set
\begin{equation}
\mathcal{L}_{\rm aed} =  -\sum_{\mathbf{(x,y})} \log p(y_u|y_{1:u-1}, x_{1:T}). \label{equ:offline}
\end{equation}

In the streaming setting, the model generates predictions at timestamps denoted by $a=(t_1, \cdots, t_u, \cdots, t_U)$, rather than waiting to the end of an utterance, where $t_u \leq t_{u+1}$ and $0<t_u\leq T$. We call the prediction timestamp sequence as an alignment $a$ between speech $x_{1:T}$ and token labels $y_{1:U}$. 
$\mathcal{A}_T^U=\{a\}$ denotes all alignments between $x_{1:T}$ and $y_{1:U}$.
The streaming model parameter $\theta_{\rm s}$ is optimized through  
\begin{equation}
    \min_{\theta_{\rm s}} \sum_{(\mathbf{x,y})}\sum_{a \in \mathcal{A}_T^U}\sum_{u=1}^{U} -\log p(y_u|y_{1:u-1}, x_{1:t_u}).  \label{equ:stream}
\end{equation}
The offline modeling can be considered a special case of streaming modeling, i.e., the alignment is unique with all $t_u=T$. 
The following two subsections briefly describe two modeling methods used in our hybrid approach.

\subsection{Attention based encoder decoder }\label{sec:subsec:aed_streaming}

 AED consists of an encoder, a decoder, and attention modules, which connect corresponding layers in the encoder and decoder as demonstrated in Figure~\ref{fig:ead}.
 The encoder generates the context representation $h_{1:T}$ from input $x_{1:T}$ \footnote{A down-sampling module might be applied in the speech encoder. For simplicity, we still use $T$ as the encoder output sequence length.}.
 The decoder state $s^l_u$ is estimated based on previous states and encoder outputs
 \begin{equation}
 s^l_u = f_{\theta_{\rm dec}}(h_{1:T}, s^l_{1:u-1}, y_{u-1}), 
 \end{equation}
 where $f_{\theta_{\rm dec}}$ is the neural network parameterized with $\theta_{\rm dec}$ and $l\in[1,L]$ is the layer index.
 
 When the AED is extended to the streaming applications~\citep{Raffel2017OnlineAL,Arivazhagan2019MonotonicIL}, a critical question has been raised: \\
 \textit{how do we decide the write/read
strategy for the decoder?} 

Assuming the AED model is Transformer based, and tokens $y_{1:u-1}$ have been decoded before timestep $t$ during inference.
The next AED decoder state $s_u^l(t)$ is associated with partial speech encoder outputs $h_{1:t}$ as well as a partial alignment $a' \in \mathcal{A}_{t}^{u-1}$ between $h_{1:t}$ and $y_{1:u-1}$. 
The computation of a Transformer decoder layer~\citep{Vaswani2017AttentionIA} includes a self-attention module and a cross-attention module. The self-attention module models the relevant information from previous decoder states 
\begin{equation}
\hat{s}^l_{{a'}} = [s^l_1({t}_1), \cdots, s^l_{u-1}({t}_{u-1})], 
\end{equation}
where ${t}_{u-1}$ is the prediction timestamp for token $u-1$ in alignment ${a'}$. The cross-attention module extracts information from the encoder outputs $h_{1:t}$.
The decoder state computation is modified as
\begin{equation}\label{equ:stream_dec}
    s^l_u(t) = f_{\theta_{\rm dec}}(h_{1:t}, \hat{s}^l_{{a'}}, y_{u-1}).
\end{equation}
To cover all read/write paths during training,  we need to enumerate all possible alignments at every timestep given the output token sequence $y_{1:U}$.
The alignment numbers would be $O(\frac{T'!(T'-U)!}{U!})$ and it is prohibitively expensive. 
In AED based methods, such as Monotonic Infinite Lookback Attention (MILk)~\citep{Arivazhagan2019MonotonicIL} and Monotonic Multihead Attention (MMA)~\citep{xma2020monotonic}, an estimation of context vector is used to avoid enumerating alignments. 
In Cross Attention Augmented Transducer (CAAT)~\citep{Liu2021CrossAA}, the self-attention modules in the joiner are dropped to decouple $y_{1:u-1}$ and $h_{1:t}$.

\subsection{Transducer}\label{sec:transducer}
 
 A Transducer has three main components. A speech encoder $\theta_{\rm enc}$ forms the context speech representation $h_{1:T}$ from speech input 
 $x_{1:T}$, a predictor $\theta_{\rm pred}$ models the linguistic information conditioned on previous 
 target tokens, and a joiner $\theta_{\rm joiner}$ merges acoustic and linguistic representations to predict outputs for every speech input feature, as shown in \autoref{fig:transducer}.
 The encoder and predictor are usually modeled with a recurrent neural network (RNN)~\citep{Graves2012SequenceTW} or Transformer~\citep{Zhang2020TransformerTA} architecture.
The joiner module is a feed-forward network which expands input from speech encoder $h_t$ and predictor output $s^L_u$ to a $T \times U$ matrix with component $z(t,u)$:
\begin{equation}
    z(t,u) = f_{\theta_{\rm joiner}}(h_t, s^L_u). 
 \label{equ:transducer_joiner}
\end{equation}

A linear projection $W^{\rm out}\in \mathcal{R}^{d\times|\mathcal{V}\cup \varnothing|}$ is applied to $z(t,u)$ to obtain logits for every output token $k\in \mathcal{V}\cup{\varnothing}$. 
 A blank token $\varnothing$ is generated if there is no good match between non-blank tokens and current $h_t$.
The RNN-T loss is optimized using the forward-backward algorithm:
\begin{equation}
\begin{split}
\alpha_{t, u} =   \mathrm{LA}\Bigl(&\alpha_{t,u-1} + \log p\bigl(y_u|z(t,u-1)\bigr), \\
& \alpha_{t-1, u} + \log p\bigl(\varnothing|z(t,u)\bigr) \Bigr)   , 
\end{split}
\end{equation}
\begin{equation}
    \mathcal{L}_{\rm rnn-t} = -\alpha_{T, U} - \log p( \varnothing|T, U),
\end{equation}
where $\mathrm{LA}(x,y)=\log(\exp^x+\exp^y)$ and $\alpha_{0,\phi}$ is initialized as 0.
Transducer is well suited to the streaming task since it can learn read/write policy from data implicitly, i.e., 
a blank token indicates a read operation and a non-blank token indicates a write operation. 

\section{Methods}
In this study, we choose the Transformer-Transducer (T-T)~\citep{Zhang2020TransformerTA} as the backbone in the proposed TAED system. 
For the streaming setting, the speech encoder is based on the chunkwise implementation~\citep{Chiu2017MonotonicCA,Chen2020DevelopingRS}, which receives and computes 
new speech input data by chunk size $N$ instead of one frame each time.

\subsection{TAED}
 TAED combines both Transducer and AED into one model, as illustrated in Figure~\ref{fig:hybrid_TAED}. 
 The speech Transformer encoder is shared between Transducer and AED models. 
The predictor in Transducer is replaced by the AED decoder. Outputs of the new predictor are results of both speech encoder outputs and predicted tokens, hence they are more informative for the joiner.

Transducer and AED models are optimized together with two criteria, RNN-T loss for the Transducer's joiner outputs and cross entropy loss for the AED decoder outputs. The overall loss $\mathcal{L}_{taed}$ is summation of two losses
\begin{equation}
    \mathcal{L}_{\rm tead} = \mathcal{L}_{\rm rnn-t} + \mathcal{L}_{\rm aed}. \label{equ:two_criteria}
\end{equation}
The model is evaluated based on the outputs from the Transducer's joiner.

\subsection{Transducer optimization with chunk- based RNN-T synchronization scheme}\label{sec:chunk_encoder}
When we attempt to extend TAED to the streaming scenario, we encounter the same streaming read/write issue discussed in \cref{sec:subsec:aed_streaming}.
In order to avoid enumerating exponential increased alignments, we adopt a different approach and 
 modify the inference logic to match the training and inference conditions. 

In the conventional streaming decoder inference, when the new speech encoder output $h_{t}$ is available, the new decoder state $s^l_u(t)$ is estimated via $h_{1:t}$ and previous computed decoder states $\hat{s}^l_{a'}$, which are based on $h_{1:t'}$ and $t' \le t$, as shown in Eq.~\eqref{equ:stream_dec}.  
In the proposed solution, we update all previous decoder states given  speech encoder outputs $h_{1:t}$, and $\hat{s}^l_{a'}$ is replaced by $\hat{s}^l_{a(t)}$, 
\begin{equation}
    \hat{s}^l_{a(t)} = [s_1^l(t), \cdots, s_{u-1}^l(t)],
\end{equation}
where ${a(t)}$ stands for a special alignment where all tokens are aligned to timestamp $t_{u}=t$. 
\begin{figure}
    \centering
\includegraphics[width=0.95\columnwidth]{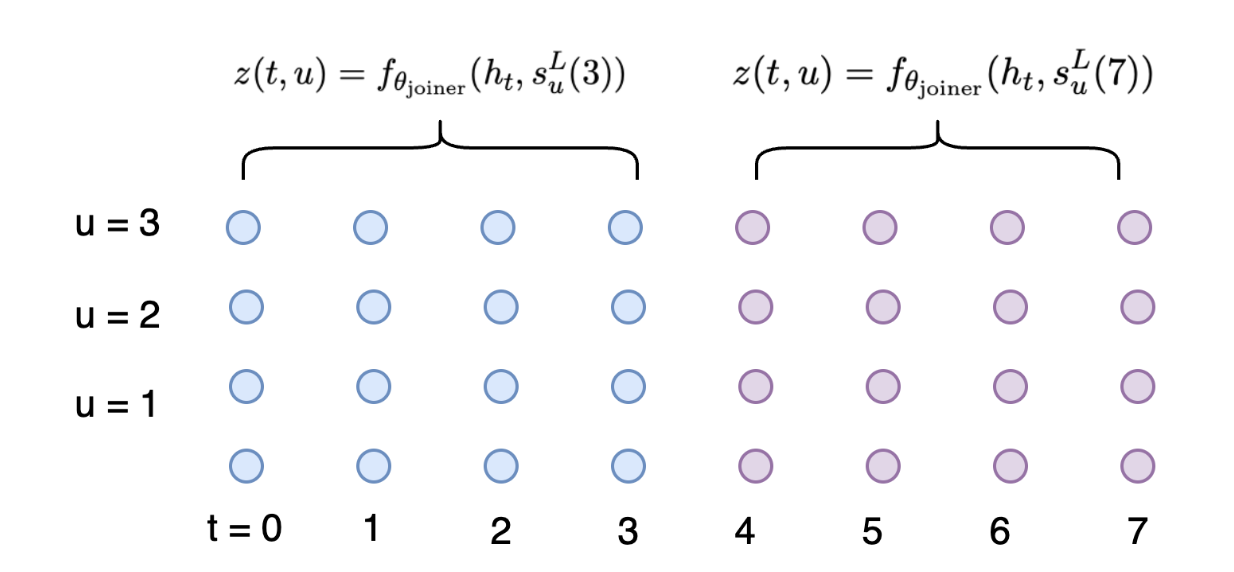}
    \caption{Chunk-based RNN-T synchronization. Frames from $t\in[0,3]$ are from the first chunk (\textcolor{cyan}{cyan}) and $t\in[4,7]$ belong to the second chunk (\textcolor{violet}{violet}).}
    \label{fig:chunk_RNNT_synchronization}
\end{figure}

There are two reasons behind this modification.
First, we expect the state representation would be more accurate if all decoder states are updated when more speech data is available.   
Second, the modification helps to reduce the huge number of alignments between $y_{1:U}$ and $h_{1:t}$ to one during training, i.e., $a(t)$.
Compared with the conventional AED training, it only increases the decoder forward computation by $T$ times. 

The computation is further reduced when the chunk-based encoder is used. 
Given two decoder states $s_u^l(t)$ and $s_u^l(\delta(t))$, where $\delta(t)$ is the last frame index of the chunk which frame $t$ belongs to and $t \leq \delta(t)$, 
$s_u^l(\delta(t))$ is more informative than $s_u^l(t)$ since the former is exposed to more speech input. During inference, $s_u^l(t)$ and $s_u^l(\delta(t))$ are available at the same time when the speech chunk data is available.  
Therefore, we replace all $s_u^l(t)$ with corresponding $s_u^l(\delta(t))$ for both inference and training.
If $N$ is the number of speech encoder output frames from one chunk of speech input,
chunk-based computation helps to reduce the decoder computation cost by $N$ times during training, since we only need to update the decoder states every $N$ frames instead of every frame.  In summary, the computation of $s^l_u(t)$ is modified from Eq.~\eqref{equ:stream_dec} as below
\begin{equation}
    s^{l}_u(t) = f_{\theta_{\rm dec}}(  h_{1:\delta(t)}, \hat{s}^l_{a(\delta(t))}, y_{u-1}),
\end{equation} \label{equ:chunk_training}
and the joiner output $z(t,u)$ in Eq.~\eqref{equ:transducer_joiner} is updated as
\begin{equation}
    z(t,u) = f_{\theta_{\rm joiner}}(h_t, s^L_u(t)).
\end{equation}
The chunk-based RNN-T synchronization is depicted in \autoref{fig:chunk_RNNT_synchronization}.
The number of speech encoder output frames in one chunk is 4. $z(t,u)$s in the first chunk (cyan) are calculated with $s_u^L(3)$ and the second chunk (violet) is based on $s_u^L(7)$. 

When the chunk size $N$ is longer than the utterance length $T$, the chunk-based TAED streaming training is the same as the offline training with similar computation cost as Transducer.

\subsection{AED optimization with fast alignment}\label{sec:aed_alignment}
Besides optimizing the model with the RNN-T loss aforementioned, we also include another auxiliary loss $\mathcal{L}_{\rm aed}$ for the AED model, as shown in Eq.~\eqref{equ:two_criteria}.
A straightforward approach is to optimize the AED modules as an offline AED model as Eq.~\eqref{equ:offline}.
However, an offline AED model could lead to high latency if it is used for streaming applications. 
Hence, we also introduce a simple ``streaming''-mode AED training by creating an even alignment $a^e$ of the target tokens against the speech encoder outputs, i.e., 
$t_u^e=\lfloor u*T'/U \rfloor$, where $\lfloor x \rfloor$ is the floor operation on $x$.  Furthermore, we can manipulate the alignment pace with an alignment speedup factor $\lambda > 0$, and the new alignment $a^\lambda$ is with timestep 
$t_u^{\lambda} = \max(T, \lfloor \frac{u*T'}{{U*\lambda}} \rfloor)$. 
When $\lambda>1.0$, the streaming AED model is trained with a fast alignment and is encouraged to predict new tokens with less speech data.
On the other hand, if $\lambda<\frac{1}{U}$, then $t_u^\lambda=T$ and it is equivalent to the offline training.  
The  auxiliary AED task is optimized via
\begin{equation}
    \mathcal{L}_{\rm aed}= - \sum_{(\mathbf{x,y})} \sum_{u} \log p(y_u|y_{1:u-1}, h_{1:t_u^{\lambda}}).  \label{equ:aed_even}
\end{equation}
Note an accurate alignment is not required in this approach, which could be difficult to obtain in translation-related applications.

\subsection{Blank penalty during inference}
The ratio between the number of input speech frames and the number of target tokens could be varied due to many factors, such as different speech ratios or target language units. 
In the meantime, the prediction of a blank token $\varnothing$ indicates a read operation, and a non-blank token represents a write operation, as discussed in \cref{sec:transducer}. 
During inference,  a blank penalty $\tau$ is introduced to adjust the target token fertility rate by penalizing the blank token $\varnothing$ emission probability. 
It acts as word insertion penalty used in ASR~\citep{Takeda1998EstimatingEO}:
\begin{equation}
    \hat{e}(t,u)[i_\varnothing] = e(t,u)[i_\varnothing] - \tau,
\end{equation}
where ${e}(t,u)=\textsc{LogSoftmax}_u(W^{\rm out}z(t,u))$ and $i_\varnothing$ is the index for the blank token $\varnothing$.

\begin{table*}
\centering
\small
\begin{tabular}{lcccc}
\toprule
\textbf{Method} & Synchronization & Merge Module & R/W decision & Training/Inference  \\
\midrule
 CIF~\citep{Dong2020CIFCI} & Async & Decoder & $h_{\leq j}$ & sampling+scaling/sampling \\
 HMA~\citep{Raffel2017OnlineAL} & Async & Decoder & $h_j,s_{\le i}$ & expectation/sampling \\
 MILk~\citep{Arivazhagan2019MonotonicIL} & Async & Decoder & $h_{\leq j},s_{<i}$ & expectation/sampling \\
 MMA~\citep{xma2020monotonic} & Async & Decoder & $h_{\leq j},s_{<i}$ & expectation/sampling \\
\midrule
\midrule
 CTC~\citep{Graves2006ConnectionistTC} & Sync & None & $h_{j}$ & all paths/sampling \\
 Transducer~\citep{Graves2012SequenceTW} & Sync & Joiner & $h_{j}, s_{k<i}$ & all paths/sampling \\
 CAAT~\citep{Liu2021CrossAA} & Sync & Joiner & $h_{\leq j}, s_{<i}$ & all paths/sampling \\
 TAED (this work) & Sync & Predictor, Joiner & $h_{\leq j}, s_{<i}$ & all paths/sampling \\
 \bottomrule
 \end{tabular}
 \vspace{-2mm}
\caption{Comparison of different streaming methods. ``R/W decision'' column lists information needed for R/W decision.}
\label{tab:com_streaming}
\end{table*}

\section{Comparison of streaming algorithms}

When to read new input and write new output is a fundamental question for the streaming algorithm.
Based on the choices of streaming read/write policies, they can roughly be separated into two families: pre-fixed and adaptive. The pre-fixed policy, such as Wait-$k$~\citep{Ma2019STACLST},  adopts a fixed scheme to read new input and write new output. 
On the other hand, the adaptive algorithms choose read/write policies dynamically based on the input speech data presented. The adaptive algorithms could be further separated into two categories based on input and output synchronization. 

The first category of adaptive streaming algorithms is based on the AED framework, including hard monotonic attention (HMA)~\citep{Raffel2017OnlineAL}, MILk~\citep{Arivazhagan2019MonotonicIL}, MoChA~\citep{Chiu2017MonotonicCA}, MMA~\citep{xma2020monotonic} and continuous integrate-and-fire (CIF)~\citep{Dong2020CIFCI,Chang2022ExploringCI}. Those methods extract acoustic information from the encoder outputs via attention between the encoder and decoder. The acoustic information is fused with linguistic information, which is estimated from the decoded token history, within the decoder. 
There is no explicit alignment between the input and output sequence; in other words, the outputs are \textbf{asynchronized} for the inputs.
As discussed in \cref{sec:subsec:aed_streaming}, AED models don't fit the streaming application easily, and approximations have been taken during training. For example, the alignment-dependent context vector extracted via attention between the encoder and decoder is usually replaced by a context vector expectation from alignments. 
It differs from inference, which is based on a specific alignment path sampled during decoding. 
Hence a training and inference discrepancy is inevitable, potentially hurting the streaming performance. 

The second category of adaptive streaming methods is with \textbf{synchronized} inputs and outputs, in which every output token is associated with a speech input frame. This includes CTC, Transducer, CAAT, and the proposed TAED. They combine acoustic and linguistic information within the joiner if linguistic modeling is applied. Specific read/write decisions are not required during training. This considers all alignments and is optimized via CTC loss or RNN-T loss. Hence, there is no training and inference discrepancy. The detailed comparison of different methods is listed in Table~\ref{tab:com_streaming}.

\section{Experiments}
\subsection{Experimental setup}\label{sec:exp_setting}
\noindent\textbf{Data} 
Experiments are conducted on two \textsc{MuST-C}~\citep{Gangi2019MuSTCAM}
 language pairs: English to German (EN$\rightarrow$DE) and English to Spanish (EN$\rightarrow$ES). Sequence level knowledge 
 distillation~\citep{Kim2016SequenceLevelKD} is applied to boost the ST quality~\citep{Liu2021CrossAA}. The English portion of data in the EN$\rightarrow$ES direction is used for English ASR  development and evaluation.
 The models are developed on 
 the \texttt{dev} set, and the final results are reported on the \texttt{tst-COMMON} set. 
 We also report \textsc{Librispeech}~\cite{Panayotov2015LibrispeechAA} ASR results in Appendix~\ref{sec:apd_librispeech} for convenient comparison with other ASR systems.
 
 \noindent\textbf{Evaluation} The ASR quality is measured with word error rate (WER), and the ST quality is reported by  
case-sensitive detokenized BLEU, which is based on the default \textsc{sacrebleu} options~\citep{post-2018-call}\footnote{case.mixed+numrefs.?+smooth.exp+tok.none+version.1.5.1}. 
 Latency is measured with Average Lagging (AL)~\citep{Ma2019STACLST} using SimualEval~\citep{ma-etal-2020-simuleval}.

\noindent\textbf{Model configuration} Input speech is represented as 80-dimensional log mel-filterbank coefficients computed every 10ms with a 25ms window. Global channel mean and variance normalization is applied. The SpecAugment~\citep{park2019specaugment} data augmentation with the LB policy is applied in all experiments.
The target vocabulary consists of 1000 ``unigram'' subword units learned by SentencePiece~\citep{Kudo2018SentencePieceAS} with full character coverage of all training text data.  

We choose the Transformer-Transducer (T-T)~\citep{Zhang2020TransformerTA} as our Transducer baseline model. The speech encoder starts with two 
casual convolution layers with a kernel size of three and a stride size of two. The input speech features are down-sampled by four and then processed by 16 
chunk-wise Transformer layers with relative positional embedding~\citep{Shaw2018SelfAttentionWR}. For the streaming case, the speech encoder can access speech data in all chunks before and one chunk ahead of the current timestep~\citep{Wu2020StreamingTA,Shi2020EmformerEM,Liu2021CrossAA}. We sweep over chunk size from 160ms to 640ms. For the offline model, we simply set a chunk size larger than any utterance to be processed as discussed in \cref{sec:chunk_encoder}.
There are two Transformer layers in the predictor module.  
The Transformer layers in both the speech encoder and predictor have an input embedding size of 512, 8 
attention heads, and middle layer dimension 2048. The joiner module is a feed-forward neural network as 
T-T~\citep{Zhang2020TransformerTA}. The TAED follows the same configuration as the T-T baseline, except the predictor module is replaced by an AED decoder with extra attention modules to connect the outputs from the speech encoder. The total number of parameters is approximately 59M for both Transducer and TAED configurations.

\begin{table}
    \centering
    \small
    \begin{tabular}{ccc}
    \toprule
    \multirow{2}{*}{Model} & \multicolumn{2}{c}{BLEU ($\uparrow$)} \\ \cmidrule(lr){2-3}
     & EN$\rightarrow$DE & EN$\rightarrow$ES  \\
    \midrule
       AED~\citep{Wang2020FairseqSF} & 22.7 & 27.2 \\
       AED~\citep{Inaguma2020ESPnetSTAS} & 22.9 & 28.0 \\
       CAAT~\citep{Liu2021CrossAA} & 23.1 & 27.6  \\
    \midrule
       Transducer & 24.9 & 28.0 \\
       TAED & \bf{25.7} & \bf{29.6} \\
    \bottomrule
    \end{tabular}
    \vspace{-2mm}
    \caption{Comparison of offline ST on the \textsc{MuST-C} \texttt{tst-COMMON} set.}
    \label{tab:Ofl_translation}
    \vspace{-2mm}
\end{table}

\noindent\textbf{Hyper-parameter setting} The model is pre-trained with the ASR task using the T-T architecture. The trained speech encoder is used to initialize the TAED models and the T-T based ST model.  
The models are fine-tuned up to 300k updates using 16 A100 GPUs. The batch size is 16k speech frames per GPU. It takes approximately one day to train the offline model and three days for the streaming model due to the overhead of the lookahead chunk and chunk-based synchronization scheme. 
Early stopping is adopted if the training makes no progress for 20 epochs.
The RAdam optimizer~\citep{Liu2019OnTV} with a learning rate 3e-4 is employed in all experiments. Label smoothing and dropout rate are both set to 0.1. 
We choose blank penalty $\tau$ 
by grid search within [0, 4.0] with step=$0.5$ on the \texttt{dev} set.
The models are trained with \textsc{Fairseq}~\citep{Wang2020FairseqSF}. 
The best ten checkpoints are averaged for inference with greedy search (beam size=1). 

\subsection{Offline results}
The results for the offline models are listed in Table~\ref{tab:Ofl_translation} and Table~\ref{tab:Ofl_asr}. In Table~\ref{tab:Ofl_translation}, our models are compared with systems reported using \textsc{MuST-C} data only. 
The first two rows are based on the AED framework, and the third one is the results from CAAT, which is the backbone in the IWSLT2021~\citep{Anastasopoulos2021FINDINGSOT} and IWSLT2022~\citep{anastasopoulos2022findings} streaming winning systems. Our Transducer baseline achieves competitive results and is comparable with 
the three systems listed above. The quality improves by 0.8 to 1.6 BLEU after we switch to the proposed TAED framework. 
Table~\ref{tab:Ofl_asr} demonstrates the corresponding ASR quality, and TAED achieves 14\% relative WER reduction compared with 
the Transducer baseline on the \texttt{tst-COMMON} set.

The results indicate Transducer can achieve competitive results with the AED based model in the ST task. 
A predictor conditioned with speech encoder outputs could provide a more accurate representation for the joiner.
The TAED can take advantage of both the Transducer and AED and achieve better results. 

\begin{table}
    \centering
    \small
    \begin{tabular}{ccc}
    \toprule
    \multirow{2}{*}{Model} & \multicolumn{2}{c}{WER ($\downarrow$)} \\  \cmidrule(lr){2-3}
    & \texttt{dev} & \texttt{tst-COMMON}  \\
    \midrule
    Transducer & 14.3 & 12.7 \\
    TAED & \bf{11.9} & \bf{10.9} \\
    \bottomrule
    \end{tabular}
    \vspace{-2mm}
    \caption{Comparison of offline ASR on the \textsc{MuST-C} \texttt{dev} and \texttt{tst-COMMON} sets.}
    \label{tab:Ofl_asr}
    \vspace{-2mm}
\end{table}

In the next experiment, we compare the impact of the AED task weight for the offline model. In Eq.~\eqref{equ:two_criteria}, the RNN-T loss and AED cross entropy loss are added to form the overall loss during training.
In Table~\ref{tab:Ofl_aed_wts}, we vary the AED task weight during training from 0.0 to 2.0.
The 2nd, 3rd, and 4th columns correspond to the AED task weight, ASR WER, and ST BLEU in the ``EN$\rightarrow$ES'' direction, respectively.
AED weight 0.0 indicates only RNN-T loss is used while AED weight = 1.0 is equivalent to the proposed mothed in~Eq.~\eqref{equ:two_criteria}. 
Without extra guidance from the AED task (AED weight=0.0), the models still outperform the Transducer models in both ASR and ST tasks, though the gain is halved.
When the AED task is introduced during training, i.e., AED weight is above 0,  we get comparable results for three AED weights: 0.5, 1.0, and 2.0.
This demonstrates that the AED guidance is essential, and the task weight is not very sensitive for the final results. In the following streaming experiments, we follow~Eq.~\eqref{equ:two_criteria} without changing the AED task weight.

\begin{table}
    \centering
    \small
    \begin{tabular}{cccc}
    \toprule
    Model & AED wts. & WER ($\downarrow$) & BLEU ($\uparrow$) \\
    \midrule
    Transducer & -- & 12.7 & 28.0 \\
    \midrule
    \multirow{4}{*}{  TAED} & 0.0 & 11.9 & 28.9  \\
                            & 0.5 & 10.9 & 30.1 \\
                            & 1.0 & 10.9 & 29.6 \\
                            & 2.0 & \bf{10.8} & \bf{30.2} \\
    \bottomrule
    \end{tabular}
    \vspace{-2mm}
    \caption{Comparison of the TAED models trained with different AED weights on the \textsc{MuST-C} \texttt{tst-COMMON} set. ``BLEU'' stands for
    the ST results from "EN$\rightarrow$ES" and ``WER'' column includes corresponding ASR results.}
    \label{tab:Ofl_aed_wts}
\end{table}

\subsection{Streaming results}
We first study the impact of the AED alignment speedup factor described in \cref{sec:aed_alignment} in Table~\ref{tab:ASR_AED_alignment} and Table~\ref{tab:ST_AED_alignment}.
In those experiments, the chunk size is set to 320ms.
The ASR results are presented in Table~\ref{tab:ASR_AED_alignment}.
The first row indicates the alignment speedup factor $\lambda$. ``Full'' means the AED model is trained as an offline ST model.
``1.0'' stands for the alignment created by evenly distributing tokens along the time axis. 
The streaming TAED model trained with the offline AED model (``Full'') achieves 12.7 WER with a large latency. 
We examine the decoding and find the model tends to generate the first non-blank token near the end of the input utterance.
The joiner learns to wait to generate reliable outputs at the end of utterances and tends to ignore the direct speech encoder outputs. 
When the fast AED alignment is adopted, i.e., $\lambda \ge 1.0$, the latency is reduced significantly from almost 6 seconds to less than 1 second. 
The larger $\lambda$ is, the smaller AL is. 
One surprising finding is that both WER and latency become smaller when $\lambda$ increases.
The WER improves from 14.7 to 12.5 when $\lambda$ increases from 1.0 to 1.2, slightly better than the TAED trained with the offline AED module.
We hypothesize that the joiner might achieve better results if it gets a synchronized signal from both the speech encoder and AED decoder outputs. 
When $\lambda$ is small, i.e., 1.0, AED decoder output might be lagged behind the speech encoder output when they are used to predict the next token.     

A similar observation is also found in ST as demonstrated in~Table~\ref{tab:ST_AED_alignment} that the fast AED alignment helps to reduce TAED latency, though the best BLEU are achieved when the offline AED module is used.
Compared to TAED models trained with offline AED module, the latency is reduced from 4+ seconds to less than 1.2 seconds for both translation directions, at the expense of BLEU score decreasing from 0.9 (EN$\rightarrow$ES) to 1.8 (EN$\rightarrow$DE).


\begin{table}
    \centering
    \small
    \begin{tabular}{c|c|c|c|c}
    \toprule
    $\lambda$ & Full & 1.0 & 1.2 & 1.4 \\
    \hline
     WER ($\downarrow$) & 12.7 & 14.7 & \bf{12.5} & 12.7 \\
     \hline
     AL ($\downarrow$) & 5894 & 1654 & \bf{849} & 907 \\
    \bottomrule
    \end{tabular}
    \vspace{-2mm}
    \caption{Comparison of AED alignment speedup factor impact for the streaming ASR performance on the \textsc{MuST-C EN} \texttt{tst-COMMON} set.}
    \label{tab:ASR_AED_alignment}
\end{table}

\begin{table}
    \centering
    \small
    \begin{tabular}{c|c|c|c|c}
    \toprule
    \multirow{2}{*}{$\lambda$} & \multicolumn{2}{c |}{EN$\rightarrow$ES }& \multicolumn{2}{c}{EN$\rightarrow$DE} \\
    \cline{2-5}
     & BLEU ($\uparrow$) & AL ($\downarrow$) & BLEU ($\uparrow$) & AL ($\downarrow$) \\
    \hline
    Full & \bf{28.3} & 4328 & \bf{24.1} & 4475 \\
    \hline
    1.0 & 27.1 & 1715 & 22.8 & 1611 \\
    \hline
    1.2 & 27.6 & 1228 & 22.6 & 1354 \\
    \hline
    1.4 & 27.6 & \bf{1120} & 22.3 & \bf{1208} \\
    \bottomrule
    \end{tabular}
    \vspace{-2mm}
    \caption{Comparison of AED alignment speedup factor impact for the streaming ST performance on the \textsc{MuST-C} \texttt{tst-COMMON} set. We set chunk size to 320ms for both EN$\rightarrow$ES and EN$\rightarrow$DE.}
    \label{tab:ST_AED_alignment}
\end{table}

In the following experiments, we compare the quality v.s. latency for TAED and Transducer. We build models with different latency by changing the chunk size from 160, 320, and 480 to 640 ms.
We present the WER v.s. AL curve in Figure~\ref{fig:mustc_asr}.  The dash lines are the WERs from the offline models, and the solid lines are for the streaming models.
The figure shows that the proposed TAED models achieve better WER than the corresponding Transducer model, varied from 1.2 to 2.1 absolute WER reduction, with similar latency values.  

\begin{figure}[h!]
        \vspace{-2mm}
		\centering
		\begin{tikzpicture}
		\definecolor{darkgreen}{RGB}{0, 100, 0}
		rgb(0,100,0)
		\begin{groupplot}
		[group style={group size= 1 by 1, horizontal sep=0.5cm, vertical sep=1.2 cm}, width=0.95\columnwidth, legend cell align={left}, height = 4.5cm, legend style={font=\tiny}, label style={font=\scriptsize}, tick label style={font=\scriptsize}]
		\nextgroupplot[
		ylabel=BLEU,
		y label style={yshift=-15pt},
		ymin=24.5, 
		ymax=30.5,
		xlabel=Average Lagging (ms),
		xmin=900, 
		xmax=1700,
        legend columns=3,
		legend style={legend pos=north west},
		xminorticks=false
		]
		\addplot[cyan, densely dashed] coordinates {(1,27.6) (3000,27.6)};
		\addlegendentry{CAAT (o)};
		\addplot[blue, densely dashed] coordinates {(1,28.0) (3000,28.0)};
		\addlegendentry{T-T (o)};
		\addplot[red, densely dashed] coordinates {(1,29.6) (3000,29.6)};
		\addlegendentry{TAED (o)};
		\addplot[cyan, mark=*] table [x index=0,y index=1, dashed, col sep=comma] {tables/mustc_caat_es.csv};
		\addlegendentry{CAAT};
		\addplot[blue, mark=*] table [x index=0,y index=1, dashed, col sep=comma] {tables/mustc_tt_es.csv};
		\addlegendentry{T-T};
		\addplot[red, mark=*] table [x index=0,y index=1, dashed, col sep=comma] {tables/mustc_taed_es.csv};
		\addlegendentry{TAED };
		\end{groupplot}
		\end{tikzpicture}
        \vspace{-2mm}
		\caption{BLEU ($\uparrow$) v.s. Average Lagging ($\downarrow$) on the \textsc{MuST-C} EN$\rightarrow$ES \texttt{tst-COMMON} dataset ($\lambda=1.4$)}
		\label{fig:mustc_st_es}
\end{figure}
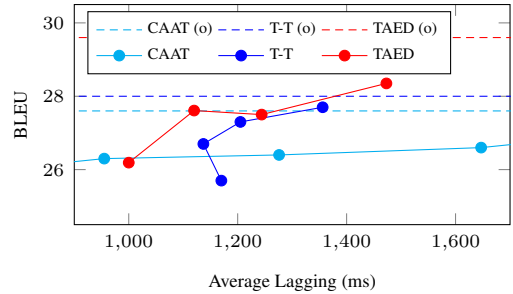

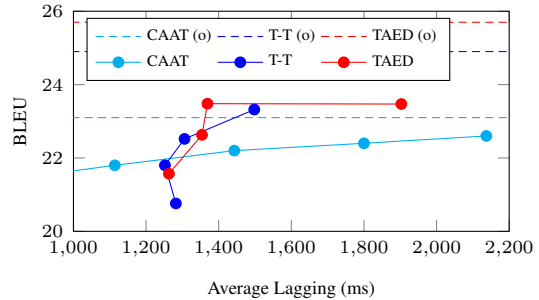
\begin{figure}[h!]
        \vspace{-2mm}
		\centering
		\begin{tikzpicture}
		\definecolor{darkgreen}{RGB}{0, 100, 0}
		rgb(0,100,0)
		\begin{groupplot}
		[group style={group size= 1 by 1, horizontal sep=0.5cm, vertical sep=1.2 cm}, width=0.95\columnwidth, legend cell align={left}, height = 4.5cm, legend style={font=\tiny}, label style={font=\scriptsize}, tick label style={font=\scriptsize}]
		\nextgroupplot[
		ylabel=BLEU,
		y label style={yshift=-15pt},
		ymin=20, 
		ymax=26,
		xlabel=Average Lagging (ms),
		xmin=1000, 
		xmax=2200,
        legend columns=3,
		legend style={legend pos=north west},
		xminorticks=false
		]
		\addplot[cyan, densely dashed] coordinates {(1,23.1) (3000,23.1)};
		\addlegendentry{CAAT (o)};
		\addplot[blue, densely dashed] coordinates {(1,24.9) (3000,24.9)};
		\addlegendentry{T-T (o)};
		\addplot[red, densely dashed] coordinates {(1,25.7) (3000,25.7)};
		\addlegendentry{TAED (o)};
		\addplot[cyan, mark=*] table [x index=0,y index=1, dashed, col sep=comma] {tables/mustc_caat_de.csv};
		\addlegendentry{CAAT};
		\addplot[blue, mark=*] table [x index=3,y index=1, dashed, col sep=comma] {tables/mustc_tt_de.csv};
		\addlegendentry{T-T};
		\addplot[red, mark=*] table [x index=3,y index=1, dashed, col sep=comma] {tables/mustc_taed_de_1.2.csv};
		\addlegendentry{TAED };
		\end{groupplot}
		\end{tikzpicture}
        \vspace{-2mm}
		\caption{BLEU ($\uparrow$) v.s. Average Lagging ($\downarrow$) on the \textsc{MuST-C} EN$\rightarrow$DE \texttt{tst-COMMON} dataset ($\lambda=1.2$)}
		\label{fig:mustc_st_de}
\end{figure}

The BLEU v.s. AL curves for ST are demonstrated in Figure~\ref{fig:mustc_st_es} and Figure~\ref{fig:mustc_st_de} for EN$\rightarrow$ES and EN$\rightarrow$DE directions, respectively. 
Besides the results from Transducer and TAED, we also include CAAT results from \citet{Liu2021CrossAA} for convenient comparison.
First, TAED consistently outperforms Transducer at different operation points in the EN$\rightarrow$ES direction and is on par with Transducer in the EN$\rightarrow$DE direction. 
We expect the TAED model outperforms the Transducer model for the EN$\rightarrow$DE direction when more latency budget is given since the offline TAED model is clearly better than the corresponding offline Transducer model.
Second, CAAT performs better at the extremely low latency region ($\sim$ 1 second AL), and TAED starts to excel CAAT when AL is beyond 1.1 seconds for EN$\rightarrow$ES and 1.3 seconds for EN$\rightarrow$DE.
TAED achieves higher BLEU scores than the offline CAAT model when the latency is more than 1.4 seconds for both directions.
The detailed results are included in Appendix~\ref{sec:streaming_results}.

\begin{figure}[h!]
		\centering
		\begin{tikzpicture}
		\definecolor{darkgreen}{RGB}{0, 100, 0}
		rgb(0,100,0)
		\begin{groupplot}
		[group style={group size= 1 by 1, horizontal sep=0.5cm, vertical sep=1.2 cm}, width=1.00\columnwidth, legend cell align={left}, height = 4.5cm, legend style={font=\tiny}, label style={font=\scriptsize}, tick label style={font=\scriptsize}]
		\nextgroupplot[
		ylabel=WER,
		y label style={yshift=-15pt},
		ymin=10, 
		ymax=17,
		xlabel=Average Lagging (ms),
		xmin=700, 
		xmax=1500,
        legend columns=2,
		legend style={legend pos=north east},
		xminorticks=false
		]
		\addplot[blue, densely dashed] coordinates {(1,12.7) (3000,12.7)};
		\addlegendentry{T-T (o)};
		\addplot[blue, mark=*] table [x index=0,y index=1, dashed, col sep=comma] {tables/mustc_tt_en.csv};
		\addlegendentry{T-T };
		\addplot[red, densely dashed] coordinates {(1,10.9) (3000,10.9)};
		\addlegendentry{TAED (o)};
		\addplot[red, mark=*] table [x index=0,y index=1, dashed, col sep=comma] {tables/mustc_taed_en.csv};
		\addlegendentry{TAED };
		\end{groupplot}
		\end{tikzpicture}
		\caption{WER ($\downarrow$) v.s. Average Lagging ($\downarrow$) on the \textsc{MuST-C} EN \texttt{tst-COMMON} dataset  ($\lambda=1.4$). ``o'' stands for the offline model.}
		\label{fig:mustc_asr}
\end{figure}
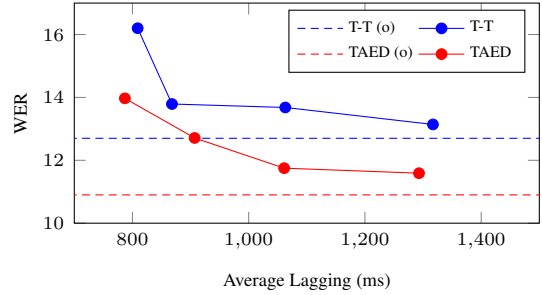

\section{Related work}
Given the advantages and weaknesses of AED and CTC/ Transducer, many works have been done to combine those methods together.  

Transducer with attention~\citep{Prabhavalkar2017ACO}, which is a Transducer variant,  also feeds the encoder outputs to the predictor. Our method is different in two aspects. 
First, we treat the TAED as a combination of two different models: Transducer and AED. 
They are optimized with equal weights during training, while Transducer with attention is optimized with RNN-T loss 
only. It is critical to achieve competitive results as shown in~Table~\ref{tab:Ofl_aed_wts}. Second, our method also includes a streaming solution while Transducer with attention can only be applied to the offline modeling.

Another solution is to combine those two methods through a two-pass approach~\citep{Watanabe2017HybridCA,Sainath2019TwoPassES,Moriya2021StreamingES}.
The first pass obtains a set of complete hypotheses using beam search.
The second pass model rescores these hypotheses by combining likelihood scores from both models and returns the result with the highest score.
An improvement along this line of research replaces the two-pass decoding with single-pass decoding, which integrates scores from CTC/Transducer with AED during the beam search~\citep{Watanabe2017HybridCA,Yan2022CTCAI}.
However, sophisticated decoding algorithms are required due to the synchronization difference between two methods. They 
also lead to high computation cost and latency~\citep{Yan2022CTCAI}.
Furthermore, the two-pass approach doesn't fit streaming applications naturally. Heuristics methods such as triggered decoding are employed~\citep{Moritz2019TriggeredAF,Moriya2021StreamingES}.  
In our proposed solution, two models are tightly integrated with native streaming support, 
and TAED predictions are synergistic results from two models.

\section{Conclusion}
In this work, we propose a new framework to integrate Transducer and AED models into one model.
The new approach ensures that the optimization covers all read/write paths and removes the discrepancy between training and evaluation for streaming applications. 
TAED achieves better results than the popular AED and Transducer modelings in ASR and ST offline tasks.
Under the streaming scenario, the TAED model consistently outperforms the Transducer baseline in both the EN ASR task and EN$\rightarrow$ES ST task while achieving comparable results in the EN$\rightarrow$DE direction.
It also excels the SOTA streaming ST system (CAAT) in medium and large latency regions.  

\section{Limitations}
The TAED model has slightly more parameters than the corresponding Transducer model due to the attention modules to connect the speech encoder and AED decoder. They have similar training time for the offline models. However, the optimization of the streaming model would require more GPU memory and computation time due to the chunk-based RNN-T synchronization scheme described in \cref{sec:subsec:aed_streaming}. In our experiments, the streaming TAED model takes about three times more training time than the offline model on the 16 A100 GPU cards, each having 40GB of GPU memory. 

In this work, we evaluate our streaming ST algorithms on two translation directions: EN$\rightarrow$ES and EN$\rightarrow$DE. The word ordering for English and Spanish languages are based on Subject-Verb-Object (SVO) while German is Subject-Object-Verb (SOV). The experiments validate the streaming algorithms on both different word ordering pair and similar word ordering pair. Our future work will extend to other source languages besides English and more language directions.

\bibliography{anthology,custom}
\bibliographystyle{acl_natbib}
\newpage
\appendix
\section{Statistics of the \textsc{MuST-C} dataset}\label{sec:must-c}
We conduct experiments on the \textsc{MuST-C}~\citep{Gangi2019MuSTCAM}.
The ASR experiments are based on the English portion of data in the EN$\rightarrow$ES direction. The ST experiments are conducted in two translation directions: EN$\rightarrow$ES and EN$\rightarrow$DE.
The detailed training data statistics are presented in Table \ref{tab:data}. The second column is the total number of hours for the speech 
training data. The third column is the number of (source) words.
\begin{table}
    \centering
    \begin{tabular}{l|c|c}
    \toprule
    \textsc{MuST-C} & hours &  \#W(m)  \\
    \hline
    $ \;\; \;\; $ EN-DE & 408 & 4.2  \\
    $ \;\; \;\; $ EN-ES & 504 & 5.2  \\
    \bottomrule
    \end{tabular}
    \caption{Data statistics for \textsc{MuST-C} training dataset. ``\#W(m)'' stands for ``number of words (million)''.}\label{tab:data}
\end{table}%

\section{Detailed streaming results}\label{sec:streaming_results}
The detailed streaming experimental results are presented in this section. We report different latency metrics from SimulEval toolkit~\citep{ma-etal-2020-simuleval}, including Average Lagging (AL)~\citep{Ma2019STACLST}, Average Proportion (AP)~\citep{Cho2016CanNM}, Differentiable Average Lagging (DAL)~\citep{Arivazhagan2019MonotonicIL}, and Length Adaptive Average Lagging (LAAL)~\citep{Papi2022OverGenerationCB}. AL, DAL and LAAL are reported with million seconds.  We report the evaluation results based on different chunk size, varied from 160, 320, 480 and 640 million seconds, from Table~\ref{tab:asr_transducer_lat} to Table~\ref{tab:st_taed_lat_de}. ``CS'' in those tables stands for chunk size. Streaming ASR results are reported as WER (Table~\ref{tab:asr_transducer_lat} and Table~\ref{tab:asr_taed_lat}). BLUE scores are reported for two translation directions in Table~\ref{tab:st_transducer_lat}, Table~\ref{tab:st_taed_lat}, Table~\ref{tab:st_transducer_lat_de} and Table~\ref{tab:st_taed_lat_de}.
\begin{table}
    \centering
    \begin{tabular}{l|c|c|c|c|c}%
    \toprule
     CS(ms) & WER & AL & LAAL & AP & DAL 
    \begin{filecontents*}{tables/mustc_tt_en.csv}
AL, WER,  LAAL, AP, DAL    ,CS
809,   16.20, 856,0.63,1177,160   
868 ,  13.79,914,0.65,1254,320   
1063,  13.68,1102,0.70,1485,480   
1317,  13.14,1350,0.71,1755,640   






\end{filecontents*}
\csvreader[head to column names]{tables/mustc_tt_en.csv}{}
    {\\\hline \CS & \WER\ & \AL & \LAAL & \AP & \DAL }
    \\
    \bottomrule
    \end{tabular}
    \caption{Transducer WER v.s. latency for different chunk sizes.}\label{tab:asr_transducer_lat}
\end{table}%

\begin{table}
    \centering
    \begin{tabular}{l|c|c|c|c|c}%
    \toprule
     CS(ms) & WER & AL & LAAL & AP & DAL 
    \begin{filecontents*}{tables/mustc_taed_en.csv}
AL, WER, LAAL, AP, DAL         ,CS
787, 13.97, 829, 0.63, 1146    ,160
907,  12.71, 947, 0.65, 1284   ,320
1061,  11.75, 1097, 0.69, 1482 ,480 
1293, 11.59, 1323, 0.74, 1742  ,640



\end{filecontents*}
\csvreader[head to column names]{tables/mustc_taed_en.csv}{}
    {\\\hline \CS & \WER\ & \AL & \LAAL & \AP & \DAL }
    \\
    \bottomrule
    \end{tabular}
    \caption{TAED WER v.s. latency for different chunk sizes($\lambda=1.4$).}\label{tab:asr_taed_lat}
\end{table}%

\begin{table}
    \centering
    \begin{tabular}{l|c|c|c|c|c}%
    \toprule
     CS(ms) & BLEU & AL & LAAL & AP & DAL 
    \begin{filecontents*}{tables/mustc_tt_es.csv}
AL, BLEU,  LAAL, AP, DAL    ,CS
1170, 25.7, 1379,0.70,1669  ,160
1137, 26.7, 1357,0.73,1679  ,320
1205, 27.3, 1433,0.74,1784  ,480
1356, 27.7, 1578,0.77,1951  ,640







\end{filecontents*}
\csvreader[head to column names]{tables/mustc_tt_es.csv}{}
    {\\\hline \CS & \BLEU\ & \AL & \LAAL & \AP & \DAL }
    \\
    \bottomrule
    \end{tabular}
    \caption{Transducer BLEU v.s. latency for different chunk sizes (EN$\rightarrow$ES).}\label{tab:st_transducer_lat}
\end{table}%

\begin{table}
    \centering
    \begin{tabular}{l|c|c|c|c|c}%
    \toprule
     CS(ms) & BLEU & AL & LAAL & AP & DAL 
    \begin{filecontents*}{tables/mustc_taed_es.csv}
AL, BLEU, LAAL, AP, DAL       ,CS 
1000, 26.19, 1224, 0.71, 1623 ,160
1120, 27.61, 1351, 0.74, 1795 ,320
1244, 27.50, 1477, 0.77, 1930 ,480
1473, 28.35, 1693, 0.79, 2188 ,640




\end{filecontents*}
\csvreader[head to column names]{tables/mustc_taed_es.csv}{}
    {\\\hline \CS & \BLEU\ & \AL & \LAAL & \AP & \DAL }
    \\
    \bottomrule
    \end{tabular}
    \caption{TAED BLEU v.s. latency for different chunk sizes (EN$\rightarrow$ES)($\lambda=1.4$).}\label{tab:st_taed_lat}
\end{table}%

\begin{table}
    \centering
    \begin{tabular}{l|c|c|c|c|c}%
    \toprule
     CS(ms) & BLEU & AL & LAAL & AP & DAL 
    \begin{filecontents*}{tables/mustc_tt_de.csv}
CS,BLEU, LAAL,AL,AP,DAL 
160,20.76,1412,1282,0.68,1618
320,21.80,1389,1252,0.70,1612
480,22.52,1447,1306,0.72,1717
640,23.32,1630,1498,0.75,1925
\end{filecontents*}
\csvreader[head to column names]{tables/mustc_tt_de.csv}{}
    {\\\hline \CS & \BLEU & \AL & \LAAL & \AP & \DAL }
    \\
    \bottomrule
    \end{tabular}
    \caption{Transducer BLEU v.s. latency for different chunk sizes (EN$\rightarrow$DE).}\label{tab:st_transducer_lat_de}
\end{table}%

\begin{table}
    \centering
    \begin{tabular}{l|c|c|c|c|c}%
    \toprule
     CS(ms) & BLEU & AL & LAAL & AP & DAL 
    \begin{filecontents*}{tables/mustc_taed_de_1.2.csv}
CS,BLEU, LAAL,AL,AP,DAL
160,21.57,1411,1263,0.72,1823
320,22.63,1530,1354,0.74,2007
480,23.48,1554,1369,0.77,2088
640,23.47,2024,1903,0.82,2597
\end{filecontents*}
\csvreader[head to column names]{tables/mustc_taed_de_1.2.csv}{}
    {\\\hline \CS & \BLEU & \AL & \LAAL & \AP & \DAL }
    \\
    \bottomrule
    \end{tabular}
    \caption{TAED BLEU v.s. latency for different chunk sizes (EN$\rightarrow$DE)($\lambda=1.2$).}\label{tab:st_taed_lat_de}
\end{table}%

\section{Librispeech ASR results}\label{sec:apd_librispeech}
The model configure is the same as \textbf{MuST-C} experiments in~\cref{sec:exp_setting}.
The models are trained with 16 A100 GPUs with batch size 20k speech frames per GPU for 300k updates.
SpecAugment~\cite{park2019specaugment} is without time warping and dropout set to 0.1. We save the checkpoints every 2500 updates and the best 10 checkpoints are averaged for the greedy search based inference. The model are
trained on the 960 hours \textbf{Librispeech}~\cite{Panayotov2015LibrispeechAA} training set and evaluated on 4 test/dev sets. 
\begin{table}
    \centering
    \begin{tabular}{l|c|c|c|c}
    \toprule
    \multirow{2}{*}{model} & \multicolumn{2}{c|}{test} & \multicolumn{2}{c}{dev} \\
    \cline{2-5}
     &  clean & other & clean & other  %
    \begin{filecontents*}{tables/librispeech_offline_streaming.csv}
model,testclean,testother,devclean,devother
Transducer(ofl),3.2,8.0,3.0,8.2
Transducer(str),4.4,11.2,4.0,11.3
TAED(ofl),3.1,7.4,2.9,7.4
TAED(str),4.2,10.4,4.1,10.8
\end{filecontents*}
\csvreader[head to column names]{tables/librispeech_offline_streaming.csv}{}%
    {\\\hline \model &  \testclean & \testother & \devclean & \devother }
    \\
    \bottomrule
    \end{tabular}
    \caption{Comparison of offline and streaming ASR on the Librispeech datasets.``ofl'' and ``str'' stand for offline and streaming moodels respectively.}
    \label{tab:librispeech_rst}
\end{table}
In~\autoref{tab:librispeech_rst}, the streaming models (``str'') are trained with chunk size equals to 320ms with one right look-ahead chunk. TAED obtains similar WERs in two clean (easy) datasets and reduces WER varied by 0.5 to 0.8 in two other (hard) datasets.


\end{document}